\documentclass[runningheads]{llncs}
\usepackage[T1]{fontenc}
\usepackage{graphicx,verbatim}
\usepackage{hyperref}
\usepackage{amsmath}
\usepackage{xcolor}
\usepackage{fontawesome}
\usepackage{marvosym} 
\usepackage{graphicx}

\begin{document}
\title{FOD-Diff: 3D Multi-Channel Patch Diffusion Model for Fiber Orientation Distribution}
\titlerunning{FOD-Diff for Fiber Orientation Distribution}
%
\begin{comment}  %% Removed for anonymized MICCAI 2025 submission
\author{First Author\inst{1}\orcidID{0000-1111-2222-3333} \and
Second Author\inst{2,3}\orcidID{1111-2222-3333-4444} \and
Third Author\inst{3}\orcidID{2222--3333-4444-5555}}
%
\authorrunning{F. Author et al.}
\institute{Princeton University, Princeton NJ 08544, USA \and
Springer Heidelberg, Tiergartenstr. 17, 69121 Heidelberg, Germany
\email{lncs@springer.com}\\
\url{http://www.springer.com/gp/computer-science/lncs} \and
ABC Institute, Rupert-Karls-University Heidelberg, Heidelberg, Germany\\
\email{\{abc,lncs\}@uni-heidelberg.de}}

\end{comment}

\author{Hao Tang\inst{1} \and
Hanyu Liu\inst{1} \and
Alessandro Perelli\inst{1} \and
Xi Chen\inst{2} \and
Chao Li\inst{1,3}\textsuperscript{\Letter}}  %% Added for anonymized MICCAI 2025 submission
\authorrunning{H. Tang et al.}
\institute{University of Dundee \and University of Bath \and University of Cambridge \\
%\email{email@anonymized.com}
}

\maketitle              % typeset the header of the contribution
\begin{abstract}
Diffusion MRI (dMRI) is a critical non-invasive technique to estimate fiber orientation distribution (FOD) for characterizing white matter integrity. Estimating FOD from single-shell low angular resolution dMRI (LAR-FOD) is limited by accuracy, whereas estimating FOD from multi-shell high angular resolution dMRI (HAR-FOD) requires a long scanning time, which limits its applicability. Diffusion models have shown promise in estimating HAR-FOD based on LAR-FOD. However, using diffusion models to efficiently generate HAR-FOD is challenging due to the large number of spherical harmonic (SH) coefficients in FOD. Here, we propose a 3D multi-channel patch diffusion model to predict HAR-FOD from LAR-FOD. We design the FOD-patch adapter by introducing the prior brain anatomy for more efficient patch-based learning. Furthermore, we introduce a voxel-level conditional coordinating module to enhance the global understanding of the model. We design the SH attention module to effectively learn the complex correlations of the SH coefficients. Our experimental results show that our method achieves the best performance in HAR-FOD prediction and outperforms other state-of-the-art methods.

\end{abstract}
\section{Introduction}
Diffusion MRI (dMRI) is the only non-invasive neuroimaging technique characterizing brain microstructure by probing water molecule diffusion~\cite{ref_article1}. Fiber bundle reconstruction based on dMRI is key to measuring fiber integrity for understanding brain physiology and pathology~\cite{ref_article1,ref_article2,ref_article3}. Traditional methods, such as diffusion tensor imaging, estimate overall fiber directions efficiently~\cite{ref_article4} but fails to model complex structures like crossings and bendings due to the single-fiber-per-voxel assumption~\cite{ref_article5}.

Advanced methods based on high angular resolution diffusion imaging\break (HARDI), offer better characterization of complex structures~\cite{ref_article2}. To obtain fiber orientation distribution (FOD), spherical deconvolution methods are developed based on the spherical harmonic (SH) function~\cite{ref_article7,ref_article8,ref_article9,ref_article10}. Despite more accurate microstructure estimation, the clinical utility of multi-shell HARDI is limited by long scanning time. In clinical practice, single-shell dMRI with lower b-values and fewer gradient directions, known as low angular resolution diffusion imaging (LARDI), is used to balance quality and efficiency~\cite{ref_article11}, which yet is limited in accurate FOD estimation. There is an unmet need to improve FOD estimation based on LARDI for wider clinical impact.
%Diffusion spectral imaging (DSI)~\cite{ref_article6}, estimate the diffusion orientation distribution function,   an indirect parameter of fiber orientation, provided intricate details but was constrained by long acquisition time. 
%improved efficiency, with q-ball imaging becoming a widely used approach~\cite{ref_article5}.

Deep learning approaches promise to meet this demand. For instance, Zeng et al. introduced FOD-Net, a 3D CNN-based model for angular super-resolution predicting multi-shell HARDI FOD (HAR-FOD) from single-shell LARDI FOD (LAR-FOD)~\cite{ref_article11}. Similarly, Silva et al. proposed a transformer-based FOD-Swin-Net~\cite{ref_con1}, and Bartlett et al. developed the spherical deconvolution network (SDNet)~\cite{ref_article12}. However, these regression-based methods face challenges in prediction accuracy and generalization~\cite{ref_article13,ref_con2}. Generative models offer high synthetic quality~\cite{ref_article13,ref_con2,ref_con3}. Methods based on generative adversarial network (GAN) are widely used in medical images~\cite{ref_article13}. For example, Jha et al. proposed a GAN-based method combining multiple attention modules to reconstruct HAR-FOD slices~\cite{ref_con2}. However, this slice-level method is limited in capturing the complex relations between voxels across the whole brain. Further, GAN-based methods have significant limitations in training stability and generation diversity  ~\cite{ref_con3,ref_article14}.
%However, this two-dimensional (2D) method is limited in capturing the complex relations between voxels, as FOD is essentially 4D, with multi-channel voxelwise connectivity over 3D image dimensions. Besides, GAN-based methods have significant limitations in training stability and generating diversity~\cite{ref_con3,ref_article14}.

Diffusion models (DMs) have recently shown their advantages in synthetic performance~\cite{ref_con3,ref_article14,ref_con4}. For instance, the denoising diffusion probability models (DDPMs) have achieved excellent performance on natural~\cite{ref_con4} and medical images~\cite{ref_article15,ref_phydiff,ref_coladiff,ref_discdiff}. Despite the success, challenges remain to efficiently optimize FOD using DMs: 1) FOD contains a large number of SH coefficients at each voxel, therefore training DMs requires large samples, which is not feasible due to acquisition costs; 2) Traditional patch learning methods based on sliding window and random patch selection are not efficient in training~\cite{ref_con5} and cannot reflect the intrinsic neuroanatomical information in FOD, which requires tailored patch training strategy; 3) FOD data essentially characterize fiber connections of the brain, indicating the integrity of anatomical structures. It is crucial to maintain the coherence of generation, which is however challenged by edge artifacts from DMs; 4) FOD encompasses multi-channel voxel-wise connectivity (SH coefficients) over 3D image dimensions. It is challenging to efficiently train the DMs to capture the complex correlations for accurate generation.

In this study, we propose a 3D Multi-Channel Diffusion Model (FOD-Diff) for HAR-FOD prediction to address these challenges. Our contributions include: (1) We design a diffusion framework to predict HAR-FOD. To mitigate the challenge of sample availability, FOD-Diff adopts a 3D patch learning strategy. As far as we know, this is the first diffusion model for FOD. (2) To further improve the efficiency of patch learning, we propose a FOD-patch adapter based on prior neuroanatomy to achieve a more efficient training strategy. (3) We introduce a voxel-level conditional coordinating module to enhance global understanding, which can enhance the coherence of the generated fibers, particularly for patch-based learning. (4) We design a spherical harmonic attention module to guide the model to effectively learn the complex correlations of the SH coefficients.

\begin{figure}[!ht]
\includegraphics[width=\textwidth]{fod-diff.png}
\caption{\textbf{Model architecture.} FPA provides the location of optimal patch $p_{k}$ by probabilistic sampling and fuses multi-scale patch feature for guidance. In the forward process, the HAR-FOD patch $S_{0}$ is gradually noised by $q(S_{t}|S_{t-1})$. In the reverse process, the condition patches are obtained by CCM and FPA from LAR-FOD to guide training. The condition patches and noised HAR-FOD patch $S_{t}$ are concatenated to train U-Net for $p_{\theta}(S_{t-1}|S_{t}, c)$. SHAM replaces the traditional output layer of U-Net to predict the forward noise patterns more effectively. In inference, predicted patches $S^{r}_{0}$ are merged by post-processing operations, and the predicted HAR-FOD is obtained.} 
\label{FOD-diff}
\end{figure}

\section{Methodology}
\subsubsection{Overview.}
Fig.~\ref{FOD-diff} illustrates the model architecture. The model aims to generate HAR-FOD using LAR-FOD as the condition. Optimal 3D multi-channel patch $p_{k}$ is selected for the diffusion process based on a sampling strategy $p_{k}\sim P(\mathbf{p}_{i})$ by incorporating anatomic tract locations in the white matter (WM) mask (see \ref{FPA}). And multi-scale anatomic information is fused and fed into the deep layer of a 3D U-Net~\cite{ref_article15} for guiding the model to efficiently adapt the FOD patch-learning (see \ref{FPA}). The selected HAR-FOD patch $S_{0}$ then undergo a Markov chain-based forward noising by $q(S_{t}|S_{t-1})$, gradually adding Gaussian noise $\epsilon$ according to the time step $t$, and generating a sequence of intermediate patches $S_{t}$ and a final approximation of isotropic Gaussian distribution $S_{T}$. 

During reverse denoising, we add multi-scale position feature as an additional condition for patch-learning by multiple Fourier frequency mappings (see \ref{CCM}). Then, the selected LAR-FOD patch together with the corresponding position patch are fed into the 3D U-Net as the conditions for the diffusion process. For each random $t$ in the training process, U-Net is trained to predict the noise pattern for obtaining $p_{\theta}(S_{t-1}|S_{t}, c)$. We design SHAM, a special output layer in U-Net to enhance the local learning of SH coefficients (see \ref{SHAM}). After training, the reverse diffusion process will generate the target patches through gradual denoising by the learned $p_{\theta}(S_{t-1}|S_{t}, c)$. The finally generated patches $S^{r}_{0}$ are merged to form the complete HAR-FOD by edge cropping, position integration based on a sliding window, and whole brain mask adjustment.

%\paragraph{Sample Heading (Fourth Level)}
%The contribution should contain no more than four %levels of headings. 

\subsection{FOD-patch Adapter}
\label{FPA}
 To more efficiently train the patch diffusion model, we design the FPA by introducing the WM mask to select suitable patches for the diffusion process. Specifically, since the valid voxels of FOD are mainly located in the white matter region, the importance of each patch in the WM mask is calculated as follows:
\begin{equation}
\textit{Imp}(\mathbf{p}_i) = \frac{\sum_{j=1}^{J_i} p_{ij}}{J_{i}}
\end{equation}
where $\mathbf{p}_{i}$ is a given patch in the same FOD, $p_{ij}$ represents the voxel value (0 or 1) in $\mathbf{p}_{i}$, and $J_{i}$ represents the total number of voxels  in the patch. Based on the $Imp$, the unnormalized probability for each patch is calculated by:
\begin{equation}
P(\mathbf{p}_i) = \left\{
\begin{array}{ll}
a\times\left((1-b) + b\times\frac{\textit{Imp}(\mathbf{p}_i)}{\max(\{\textit{Imp}(p_i)\}_{{p}_i \in \Theta})}\right), & \textit{if } \mathbf{p}_i \in \Theta_T \\
(1 - a)\times\left((1-b) + b\times\frac{\textit{Imp}(\mathbf{p}_i)}{\max(\{\textit{Imp}({p}_i)\}_{{p}_i \in \Theta})}\right), & \textit{if } \mathbf{p}_i \notin \Theta_T
\end{array}
\right.
\label{eq:P()}
\end{equation}
where $\Theta$ is the set of all patches in the original WM mask, and $\Theta_T$ is the subset of $\Theta$, which includes all the target patches in the sliding window method during the test phase. $a$ and $b$ are hyper-parameters to dynamically adjust the sampling state during training. To balance training efficiency and performance, $a$ and  $b$ are initially set as 0.99 and 0.8,  respectively, and will decrease 
%to 0.9 and 0.6,
in linear decay as training progresses. Hence, the model can focus more on the target patches with higher fiber density while maintaining the generalization of patch learning. This patch sampling mechanism is achieved by the following formula:
\begin{equation}
{p_k} \sim \left\{P_n(\mathbf{p}_i) = \frac{P(\mathbf{p}_i)}{\sum_{p_i \in \Theta} P(p_i)}\right\}
\end{equation}
where $\sim$ indicates random sampling according to the normalized probability $P_n(\mathbf{p}_i)$. $p_{k}$ represents the sampled result. After sampling the WM mask patch $p_{k}$, the module will pass the location of $p_{k}$ to the diffusion process for patch selection (Fig.~\ref{FOD-diff}).

Meanwhile, for guiding the model to adapt to the FOD patch-leaning strategy efficiently while relieving the loss of anatomic information caused by multi-layer feature extractions and a large number of LAR-FOD channels, FPA fuses $p_{k}$ with the original WM mask $M_0$ and introduces the fused feature $\mathbf{F}_{m}$ into the deeper layer of the U-Net instead of directly merging them as traditional conditions with the noise $S^{r}_{t}$ (Fig.~\ref{FOD-diff}). Specifically, $p_{k}$ and $M_0$ are processed through a series of feature extraction operations $f_{e}(\cdot)$, including a convolution layer for expanding channel dimension, an adaptive pooling, and a nonlinearity layer. The overall fusion process is as follows:
\begin{equation}
\begin{array}{ll}
\mathbf{F}_{m} = \text{ReLU}(\text{Concat}[f_{e}(p_k), f_{e}(\Theta)]) \\
\mathbf{F}_{a} = \text{Attention}(\text{3DConv}(\text{Concat}[x, \mathbf{F}_m]))
\end{array}
\end{equation}
where $Attention$ is the attention module~\cite{ref_article15} used to learn $\mathbf{F}_{m}$. $x$ is the output feature of the second downsampling layer. $\mathbf{F}_{a}$ is the fused attention feature by the attention module in the third layer of U-Net (Fig.~\ref{FOD-diff}). $Concat[\cdot\,,\cdot]$ denotes the concatenation operation. This mechanism fuses multi-scale anatomic information to make the model efficiently adapt to this patch-based learning.

\subsection{Voxel-level Conditional Coordinating Module}
\label{CCM}
Patch-based learning can enhance the learning of details, but could be limited in understanding the whole-brain FOD distribution and the patch-wise relationships. Therefore, we introduce a voxel-level conditional coordinating module (CCM) based on Patch-Diffusion~\cite{ref_con6} and COCO-GAN~\cite{ref_con7}. CCM adds additional positional channels for each voxel of the whole-brain FOD to provide structural information. Unlike simple coordinate information, CCM provides multi-scale position coding information based on Fourier transform. Specifically, after obtaining the normalized voxel position coordinates of the conditions, a frequency band is designed using logarithmic sampling to obtain different frequencies with a large span. The frequencies are calculated by:
\begin{equation}
\omega_l = 2^{\frac{l \times \text{$f_{\max}$}}{L}}, \quad l = 0, 1, \ldots, L
\end{equation}
where $L$ is the number of frequencies and $f_{\max}$ is the maximum frequency. $L$ and $f_{\max}$ are both set to 2 for balancing performance. The original coordinates are then mapped to this frequency band. The final voxel-level positional feature $\mathbf{F}_{vp}$ is expressed as follows:
\begin{equation}
\begin{array}{ll}
\mathbf{F}_p(x) = \text{Concat}[\sin(\omega_0 x), \cos(\omega_0 x), \ldots, \sin(\omega_L x), \cos(\omega_L x)] \\
\mathbf{F}_{vp} = \text{Concat}[\mathbf{F}_p(x), \mathbf{F}_p(y), \mathbf{F}_p(z)]
\end{array}
\end{equation}
where $\mathbf{F}_{p}(\cdot)$ represents the positional feature of a dimension, including x, y, and z. Finally, the patches of $\mathbf{F}_{vp}$ are selected and introduced into the U-Net (Fig.~\ref{FOD-diff}).

\subsection{Spherical Harmonic Attention Module}
\label{SHAM}
To improve the understanding of the large number of voxel-wise SH coefficients, we design the spherical harmonic attention module (SHAM). For balancing the performance, SHAM first performs global average pooling and maximum pooling on the input feature to extract the global features and local salient features of channels respectively. Then, each pooled feature is transformed and activated to obtain SH features by the SH feature extraction layer (SHFE), which are designed based on the coefficient distribution of FOD. 

In essence, the FOD channels are SH coefficients, which are obtained on SH basis, and for each voxel $v$, reconstructed FOD is expressed as follows:
\begin{equation}
F(v) = \sum_{h=0}^{h_{\max}} \sum_{m=-h}^{h} c^{h}_{m} Y^{h}_{m}(v) \label{eq:FOD}
\end{equation}
where $h_{\max}$ is the maximum harmonic order, $c^{h}_{m}$ represents the SH coefficients, $Y^{h}_{m}$ represents the SH basis function, and reconstructed FOD is their linear combination. Estimating FOD is essentially estimating SH coefficients. Further, due to the symmetry of dMRI data, FOD is represented by only even orders, so the number of orders is $1+h_{\max}/2$. Therefore, when $h_{\max}$ takes the default value of 8, the even orders of SH coefficients obtained are 0, 2, 4, 6, and 8, respectively, and the corresponding number of coefficients is 1, 5, 9, 13, 17, where different orders represent different angular frequencies. For each $h$, $2h+1$ SH functions and coefficients together form the complete basis at this frequency. Therefore, the coefficients of each order have different distribution characteristics and the relationship between the $2h+1$ coefficients is much closer and more complex. 

To guide the model to expressively perceive this channel rule and enhance the prediction of the coefficients in each order, the SHFE includes five layers, each of which contains a convolutional layer and a nonlinearity layer. The channel transformation targets of the five layers are 1, 5, 9, 13, 17 respectively,  corresponding to the SH coefficients of different orders.

The five SH features output by the pooling and SHFE are concatenated to obtain the overall SH features of the five layers. The overall SH global features and local salient features are added and converted into the SH attention weight by a convolution layer and Sigmoid activation function. This weight is directly multiplied with the output of the original output layer (Fig.~\ref{FOD-diff}) to achieve the overall adjustment of the SH coefficients in different orders and the detailed weighting of the SH coefficients in the same order.

%\begin{theorem}
%This is a sample theorem. The run-in heading is set in %bold, while
%the following text appears in italics. Definitions, %lemmas,
%propositions, and corollaries are styled the same way.
%\end{theorem}
%
% the environments 'definition', 'lemma', 'proposition', 'corollary',
% 'remark', and 'example' are defined in the LLNCS documentclass as well.
%
%\begin{proof}
%Proofs, examples, and remarks have the initial word %in italics,
%while the following text appears in normal font.
%\end{proof}

%For citations of references, Multiple citations are %grouped
%\cite{ref_article1,ref_lncs1,ref_book1},
%\cite{ref_article1,ref_book1,ref_proc1,ref_url1}.

\section{Experiments and Results}
\subsection{Datasets and Preprocessing}
We randomly select dMRI and corresponding T1-weighted images of 20 subjects in the Human Connectome Project (HCP)~\cite{ref_article17}, where 10 subjects were for training, 2 for validation, and 8 for test. The complete dMRI data of each subject contains 288 volumes (multi-shell HARDI), including 270 volumes of b=1000, 2000, 3000 $s/mm^{2}$ with 90 gradient directions for each shell and 18 b0 volumes. Due to the wide application of low b-value with 32 gradient directions in clinical practice~\cite{ref_article11}, we subsample 32 volumes of b=1000 and 1 volume of b0 from the complete dMRI according to the HCP  protocol, to obtain single-shell LARDI.

Further, we use MSMT-CSD on the multi-shell HARDI to obtain multi-shell HARDI FOD as the target of FOD-Diff (HAR-FOD). Meanwhile, we use SSMT-CSD on the single-shell LARDI to obtain single-shell LARDI FOD as the condition of FOD-Diff (LAR-FOD). $h_{max}$ is set to the default value of 8 to balance precision and complexity so that all FOD data have 45 SH coefficients. Furthermore, we obtain the white matter mask of each subject based on the T1-weighted image and the segmentation algorithm in MRTrix~\cite{ref_article18} to guide the training of the model and evaluate the subsequent experimental results. These data are cropped out the surrounding invalid voxels before entering the model, and fill in these areas after the inference is completed for higher efficiency.

\subsection{Implementation Details}
Model hyper-parameters: the loss function is mean square error; noising way is cosine noise schedule~\cite{ref_article15} of 250 steps; Adam optimizer with a learning rate of $1\times10^{-4}$ for the first 50000 iterations and $5\times10^{-5}$ for the last 50000 iterations; batch size is 4; patch size for diffusion process is $32\times32\times32$; the target size after edge clipping in inference process is $20\times20\times20$; the step sizes of the sliding window for the training and inference processes are 2 and 20; U-Net has 4 layers with channels of 128, 256, 256, 512; a and b of FPA eventually decay to 0.8 and 0.5 (Eq.\eqref{eq:P()}). All hyper-parameters are tuned to balance the performance and efficiency on the validation dataset. The model is implemented through PyTorch and in a hardware configuration with a 16GB Intel (R) Core (Tm) i9-12900H CPU and an 8GB GeForce RTX 3070Ti Laptop GPU.

\begin{table}
\centering
\caption{Performance comparison on HCP dataset. Bold represents the best result.}
\label{tab1}
\setlength{\tabcolsep}{12pt}
\small
\begin{tabular}{ccc}
\hline
Methods & ACC of WM & ACC of whole brain\\
\hline
\begin{tabular}[c]{c}
SSMT-CSD (LAR-FOD)\end{tabular} &{0.7523$\pm$0.0256}  &{0.6640$\pm$0.0145}\\
FOD-Net &{0.8828$\pm$0.0136}  &{0.8114$\pm$0.0158}\\
FOD-SWIN-Net &{0.8780$\pm$0.0158} &{0.7987$\pm$0.0161}\\
SD-Net  &{0.8837$\pm$0.0149}  &{0.8134$\pm$0.0124}\\
{\bfseries FOD-Diff (Ours)} &{\bfseries 0.8905$\pm$0.0137} &{\bfseries 0.8177$\pm$0.0138}\\
\hline
\end{tabular}
\end{table}

\subsection{Comparisons with SOTA and baselines}
We compare our model with SSMT-CSD (LAR-FOD) and 3 state-of-the-art HAR-FOD prediction approaches, including FOD-Net~\cite{ref_article11}, FOD-SWIN-Net~\cite{ref_con1} and SD-Net~\cite{ref_article12}. For comparison, these methods are implemented using the same dataset and preprocessing.

\subsubsection{Quantitative Results.}
We use the Angular Correlation Coefficient (ACC)~\cite{ref_article11} to evaluate the FOD prediction quality of each method. Table~\ref{tab1} shows the ACC of the white matter region (WM) and the ACC of the whole brain for each method. Our proposed FOD-Diff achieves better performance than other methods.

\begin{figure}
\includegraphics[width=\textwidth]{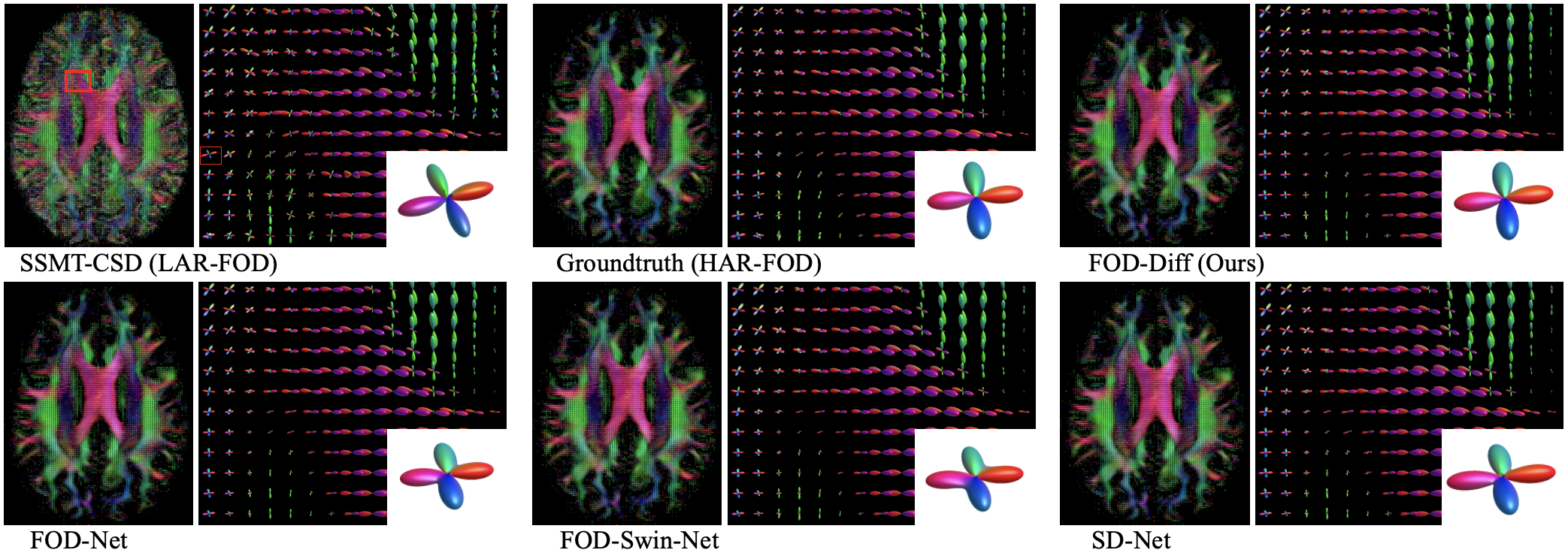}
\caption{Selected samples and their regions of interest (ROIs) for comparison. ROI is selected in the corpus callosum region and a typical cross fiber is zoomed in.} 
\label{results}
\end{figure}

\begin{table}
\centering
\caption{Ablation experiments on HCP dataset. Bold represents the best result.}
\label{tab2}
\setlength{\tabcolsep}{12pt}
%\small
\begin{tabular}{ccc}
\hline
Methods &  ACC of WM & ACC of full brain\\
\hline
FOD-Diff w/o FPA &{0.8762$\pm$0.0159} &{0.7953$\pm$0.0139}\\
FOD-Diff w/o CCM &{0.8817$\pm$0.0152} &{0.8082$\pm$0.0147}\\
FOD-Diff w/o SHAM &{0.8809$\pm$0.0143} &{0.8062$\pm$0.0136}\\
{\bfseries FOD-Diff (Ours)} &{\bfseries 0.8905$\pm$0.0137} &{\bfseries 0.8177$\pm$0.0138}\\
\hline
\end{tabular}
\end{table}

\subsubsection{Qualitative Results.}
Fig.~\ref{results} shows the selected samples of reconstructed FOD results (Eq.\eqref{eq:FOD}) from various methods. Our FOD-Diff achieves more accurate fiber distribution in the overall fiber structure and local details.

\subsection{Ablation Experiments}
To evaluate the effectiveness of each module in FOD-Diff, we test the ACC of the model ablating the modules. After removing FPA, we randomly select patches for training by the sliding window method with a step size of 5. Table~\ref{tab2} shows that ablating each module leads to decreased accuracy, supporting the effectiveness of each module. Further, FOD-Diff with  FPA removed requires approximately 150,000 iterations to converge stably during training, which shows that FPA contributes to improved training efficiency.

\section{Conclusion}
We present FOD-Diff, a 3D multi-channel patch diffusion model for FOD, with an FOD-patch adapter for efficient patch-learning, a voxel-wise conditional coordinating module for global understanding, and a spherical harmonic (SH) attention module for learning the complex correlations of SH coefficients. Our experiments show that FOD-Diff achieves the best performance in HAR-FOD synthesis, outperforming other state-of-the-art methods. As far as we know, we present the first study using diffusion models for FOD processing, providing more accurate estimates of fiber orientation while minimizing dMRI scan time.


\begin{thebibliography}{8}
\bibitem{ref_article1}
Shamir, I., Assaf, Y.: Tutorial: a guide to diffusion MRI and structural connectomics. Nature Protocols, 1-19 (2024)

\bibitem{ref_article2}
Dell'Acqua, F., Tournier, J. D.: Modelling white matter with spherical deconvolution: How and why?. NMR in Biomedicine, \textbf{32}(4), e3945 (2019)

\bibitem{ref_article3}
Jeurissen B, Descoteaux M, Mori S, Leemans A.: Diffusion MRI fiber tractography of the brain. NMR in biomedicine, \textbf{32}(4), e3785 (2019)

\bibitem{ref_article4}
Basser, P. J., Mattiello, J., LeBihan, D.: MR diffusion tensor spectroscopy and imaging. Biophysical journal, \textbf{66}(1), 259-267 (1994)

\bibitem{ref_article5}
Tuch, D. S.: Q‐ball imaging. Magnetic Resonance in Medicine, \textbf{52}(6), 1358-1372 (2004)

\bibitem{ref_article7}
Tournier, J. D., Calamante, F., Gadian, D. G., Connelly, A.: Direct estimation of the fiber orientation density function from diffusion-weighted MRI data using spherical deconvolution. Neuroimage, \textbf{23}(3), 1176-1185 (2004)

\bibitem{ref_article8}
Tournier, J. D., Calamante, F., Connelly, A.: Robust determination of the fibre orientation distribution in diffusion MRI: Non-negativity constrained super-resolved spherical deconvolution. NeuroImage, \textbf{35}(4), 1459-1472 (2007)

\bibitem{ref_article9}
Khan, W., Egorova, N., Khlif, M. S., Mito, R., Dhollander, T., Brodtmann, A.: Three-tissue compositional analysis reveals in-vivo microstructural heterogeneity of white matter hyperintensities following stroke. Neuroimage, \textbf{218}, 116869 (2020)

\bibitem{ref_article10}
Jeurissen, B., Tournier, J. D., Dhollander, T., Connelly, A., Sijbers, J.: Multi-tissue constrained spherical deconvolution for improved analysis of multi-shell diffusion MRI data. NeuroImage, \textbf{103}, 411-426 (2014)

\bibitem{ref_article11}
Zeng, R., Lv, J., Wang, H., Zhou, L., Barnett, M., Calamante, F., Wang, C.: FOD-Net: A deep learning method for fiber orientation distribution angular super resolution. Medical Image Analysis, \textbf{79}, 102431 (2022)

\bibitem{ref_con1}
Da Silva, M. O., Santana, C. P., Do Carmo, D. S., Rittner, L.: FOD-Swin-Net: angular super resolution of fiber orientation distribution using a transformer-based deep model. In: 2024 IEEE International Symposium on Biomedical Imaging (ISBI), pp. 1-5. IEEE (2024)

\bibitem{ref_article12}
Bartlett, J. J., Davey, C. E., Johnston, L. A., Duan, J.: Recovering high‐quality fiber orientation distributions from a reduced number of diffusion‐weighted images using a model‐driven deep learning architecture. Magnetic Resonance in Medicine, \textbf{92}(5), 2193–2206 (2024)

\bibitem{ref_article13}
Zhang, A., Xing, L., Zou, J., Wu, J. C.: Shifting machine learning for healthcare from development to deployment and from models to data. Nature Biomedical Engineering, \textbf{6}(12), 1330-1345 (2022)

\bibitem{ref_con2}
Jha, R.R., Gupta, H., Pathak, S.K., Joshi, A., Schneider, W., Kumar, R., Bhavsar, A., Nigam, A.: PA-GAN: Parallel Attention Based Gan for Enhancement of FODF. In: 2023 IEEE 20th International Symposium on Biomedical Imaging (ISBI), pp. 1-5. IEEE (2023)

\bibitem{ref_con3}
Dhariwal, P., Nichol, A.: Diffusion models beat gans on image synthesis. Advances in neural information processing systems \textbf{34}, 8780-8794 (2021)

\bibitem{ref_article14}
Yang, L., Zhang, Z., Song, Y., Hong, S., Xu, R., Zhao, Y., Zhang, W., Cui, B., Yang, M.: Diffusion models: A comprehensive survey of methods and applications. ACM Computing Surveys, \textbf{56}(4), 1-39 (2023)

\bibitem{ref_con4}
Ho, J., Jain, A., Abbeel, P.: Denoising diffusion probabilistic models. Advances in neural information processing systems \textbf{33}, 6840-6851 (2020)

\bibitem{ref_article15}
Dorjsembe, Z., Pao, H. K., Odonchimed, S., Xiao, F.: Conditional diffusion models for semantic 3D brain MRI synthesis. IEEE Journal of Biomedical and Health Informatics, \textbf{28}(7), pp.4084–4093 (2024)

\bibitem{ref_phydiff}
Zhang, J., Yan, R., Perelli, A., Chen, X., Li, C.: Phy-Diff: Physics-Guided Hourglass Diffusion Model for Diffusion MRI Synthesis. In International Conference on Medical Image Computing and Computer-Assisted Intervention, pp. 345-355. Springer (2024)

\bibitem{ref_coladiff}
Jiang, L., Mao, Y., Wang, X., Chen, X., Li, C: Cola-diff: Conditional latent diffusion model for multi-modal mri synthesis. In International Conference on Medical Image Computing and Computer-Assisted Intervention, pp. 398-408. Springer (2023)

\bibitem{ref_discdiff}
Mao, Y., Jiang, L., Chen, X., Li, C.: Disc-diff: Disentangled conditional diffusion model for multi-contrast mri super-resolution. In International Conference on Medical Image Computing and Computer-Assisted Intervention, pp. 387-397. Springer (2023)

\bibitem{ref_con5}
Wang, X., Price, S., Li, C.: Multi-task learning of histology and molecular markers for classifying diffuse glioma. In International Conference on Medical Image Computing and Computer-Assisted Intervention, pp. 551-561. Springer (2023)

\bibitem{ref_con6}
Wang, Z., Jiang, Y., Zheng, H., Wang, P., He, P., Wang, Z., Chen, W., Zhou, M.: Patch Diffusion: Faster and More Data-Efficient Training of Diffusion Models. Advances in Neural Information Processing Systems \textbf{36}, 72137–72154 (2023)

\bibitem{ref_con7}
Lin, C. H., Chang, C. C., Chen, Y. S., Juan, D. C., Wei, W., Chen, H. T.: Coco-gan: Generation by parts via conditional coordinating. In Proceedings of the IEEE/CVF international conference on computer vision, pp. 4512-4521 (2019)

\bibitem{ref_article17}
Van Essen, D. C., Smith, S. M., Barch, D. M., Behrens, T. E., Yacoub, E., Ugurbil, K.: The WU-Minn human connectome project: an overview. Neuroimage, \textbf{80}, 62-79 (2013)

\bibitem{ref_article18}
Tournier, J. D., Smith, R., Raffelt, D., Tabbara, R., Dhollander, T., Pietsch, M., Christiaens, D., Jeurissen, B., Yeh, C. H., Connelly, A.: MRtrix3: A fast, flexible and open software framework for medical image processing and visualisation. Neuroimage, \textbf{202}, 116137 (2019)

\end{thebibliography}
%

\end{document}